\documentclass{article}
\usepackage{spconf,amsmath,epsfig}
\usepackage{amsfonts}
\usepackage{hyperref}

\begin{document}\sloppy

\def\x{{\mathbf x}}
\def\L{{\cal L}}

\title{Feature Fusion Effects of Tensor Product Representation on (De)Compositional Network for Caption Generation for Images}
%
\name{Chiranjib Sur}
\address{chiranjibsur@gmail.com}

\maketitle

\begin{abstract}
Progress in image captioning is gradually getting complex as researchers try to generalized the model and define the representation between visual features and natural language processing. This work tried to define such kind of relationship in the form of representation called Tensor Product Representation (TPR) which generalized the scheme of language modeling and structuring the linguistic attributes (related to grammar and parts of speech of language) which will provide a much better structure and grammatically correct sentence. TPR enables better and unique representation and structuring of the feature space and will enable better sentence composition from these representations. A large part of the different ways of defining and improving these TPR are discussed and their performance with respect to the traditional procedures and feature representations are evaluated for image captioning application. The new models achieved considerable improvement than the corresponding previous architectures. 
\end{abstract}

\section{Introduction}
\label{sec:intro}
Image captioning application are required for transferability between visual features, related to images and videos, and textual contents. Recent works in image captioning with traditional visual features, where the architecture tries to evolve the lower level features of whole image into captions. However, the effectiveness saturated and in this work, we have introduced other procedures that can enhance the effectiveness of these architectures for better caption generation. 
A bidirectional multi-modal retriever was using visual features and language embedding in \cite{Karpathy2014Deep}, while \cite{wang2016image} proposed a bidirectional LSTM decoder. \cite{vinyals2017show} introduced generative model with advanced features. Gradually, attentions were introduced like adaptive regional emphasis based attention model \cite{lu2017knowing}, \cite{wu2017image} utilized a high-level concept based attribute attention layer, \cite{yang1605encode} introduced review attention for decoding, \cite{yao2017boosting} hybridized image attention with LSTM as a sequence-to-sequence model, \cite{rennie2017self} utilized self-critical sequence attention training with mixed reinforcement learning for performance improvement. Meanwhile, \cite{gan2017semantic} pioneered a Semantic Compositional Network (SCN) for sentence generation from images, with additional semantic concepts from image features. As the research moved further, \cite{anderson2018bottom} introduced bottom-up top-down mechanism with  region based image feature tensor attention from Faster R-CNN model.  \cite{lu2018neural} introduced a sentence template based approach with explicit slots correlated with different specific image regions using R-CNN objects predictors. Another instance of higher level attribute attention was introduced by \cite{chen2018show} where the attributes were the objects detected in the images using a separate RNN network was used for detection of these good objects from the images in a sequence that can be favorable for better caption generation. These model leveraged on co-occurrence  dependencies among object attributes and used an inference representation based on it. In contrary, our approach utilized only image features and emphasized on better representation and caption generation. A direct comparison of these external attention models with our approach will be unjustified, but we have reported competitive performance with our approach.

The rest of the document is arranged with 
description of TPR in Section \ref{sec:tp}, details of our architecture in Section \ref{sec:arc}, results and analysis in Section \ref{sec:res}  and conclusion in Section \ref{sec:discussion}.


\section{Tensor Product Representation} 
\label{sec:tp}
Tensor Product \cite{smolensky1990tensor} is the systematic composition of a series of tensors that can be utilized for special representation with structured interpretation and has nice algebraic properties that can retrieve nearest composite components. However, for our applications, we are dealing with some special situation of tensor products where one of them consists of orthogonal structures and thus creating the perfectly orthogonal segments of feature space to represent the data and the cumulative representation can be well utilized for inference, while the reverse multiplication of the orthogonal representation can retrieve the original space representations. While there can be different ways of generation of tensor products, we have concentrated on deterministic approaches with Hadamard matrix and due to its limitations, we switched to deterministic approximation techniques for tensor product generation.
Let we have sentence with word $w_1,\ldots,w_n$ and word embedding $\textbf{W}_e \in \mathbb{R}^{v \times e}$, we can transfer one hot vector for each word $w_i$ as $(\textbf{W}_e)_i \in \mathbb{R}^{1 \times e}$, we have, 
\begin{equation}
 \textbf{s}_{H} = \sum (\textbf{W}_e)_i  * f_j^T
\end{equation}
for $w_j$ at $i$ and $\textbf{s}_{H}$ is the TPR. 
Conversely, to retrieve the information from the TPR, for each $j \in N = \{1,\ldots,n\}$, we have,  
\begin{equation}
  (w_p)_{j} =  \textbf{s}_{H} * f_j
\end{equation}  
and if we consider nearest neighbor for $(w_p)_{j} $ in $\textbf{W}_e$, we find, 
\begin{multline}
 (w_p)_{j} = \arg \min\limits_{k} \{ (\textbf{W}_e)_k \mid \min ||(\textbf{W}_e)_k - (w_p)_{j}|| \}   \\
 = (\textbf{W}_e)_{k=i}  = (\textbf{W}_e)_{i} = w_{j}
\end{multline}
We have tested that the retrieval rate is 100\% correct for word embedding like Word2Vec, GloVe \cite{pennington2017stanford} for any dimension. The accuracy of the retrieval is not because of the dimension or the embedding, but due to the mutual orthogonal matrix $f_j$ which creates space for real $r_i$ to be segregated when $f_i$ is multiplied with $r_i f_i^T$ as  $r_i f_i^T f_i$.
Tensor Product Representation (TPR) \cite{smolensky1990tensor,palangi2017question,Sur2018Representation,huang2018tensor,huang2019tensor}  is made scalable through approximating the variations and thus deviates the tensors from being completely orthogonal and using series of non-linearity and memory network parameter estimations. However, the effectiveness of the TPR comes from the uniqueness of the feature space and the TPR itself. Even, the potential of the TPR to be able to help in generalization is immense as new representations get generated from contexts (image features) and can be said to  have the same representation that could have been generated by the corresponding caption of that image. 
Consider an image $I$, with caption $\textbf{H}$. Assume that a caption consists of $n$ words including the start of a sentence and stop of a sentence. We define $\textbf{H}_x = [(\textbf{h}_x)_1,(\textbf{h}_x)_2, \ldots ,(\textbf{h}_x)_n]$, where $(\textbf{h}_x)_t \in \mathbb{R}^V$ is a one-hot encoding vector of dimension $V$ and $V$ is the size of the vocabulary. The length $n$ usually varies from caption to caption. $\textbf{W}_e \in \mathbb{R}^{d\times V}$ is a word embedding matrix, the i-th column of which is the embedding vector of the i-th word in the vocabulary; it is obtained from GLoVe \cite{pennington2017stanford} algorithm with zero mean. 
$\tilde{\textbf{S}}_{t-1} \in \mathbb{R}^{d\times d}$ at time $t-1$ is defined as, 
\begin{equation}
 \tilde{\textbf{S}}_{t-1} = \sum\limits_{i=1}^{t-1} \textbf{W}_e (\textbf{h}_x)_i \textbf{r}^T_i = \sum\limits_{i=1}^{t-1} \textbf{x}_i \textbf{r}^T_i
\end{equation}
where $\textbf{r}_i$ is the role vector for word $\textbf{x}_i$, and $\textbf{r}_i^T$ is transpose of $\textbf{r}_i$. With an orthogonal Hadamard matrix, we have $\textbf{r}_t = \textbf{u}_t (t = 1,\ldots,n)$.
The attention vector $\textbf{a}_t^{(u)} \in \mathbb{R}^d$ is given by,
\begin{equation}
 \textbf{a}_t^{(u)} = \sigma_g (\textbf{W}_a^{(u)} \textbf{h}_{t-1} + \textbf{W}_s^{(u)} vec(\tilde{\textbf{S}}_{t-1}) + \textbf{b}_a^{(u)})
\end{equation}
where $\sigma_g(.)$ is the logistic sigmoid function; $\textbf{W}_a^{(u)} \in \mathbb{R}^{d\times 512}$, $\textbf{W}_s^{(u)} \in \mathbb{R}^{d\times 1024}$, $\textbf{b}_a^{(u)} \in \mathbb{R}^{d}$, $vec(.)$ is the vectorization operation. The feature vector $\textbf{q}_t \in \mathbb{R}^{2048}$ is given by,
\begin{equation}
 \textbf{q}_t = \textbf{v} \odot \textbf{a}_t^{(v)}
\end{equation}
where the operator $\odot$ denotes the Hadamard product. TPR $\textbf{S}_t \in \mathbb{R}^{d\times d}$ of word $\textbf{x}_t$ is obtained by,
\begin{equation}
 \textbf{S}_t = \sigma_h (\textbf{C}_s \textbf{q}_t + \textbf{B}_s)
\end{equation}
where $\sigma_h(.)$ is the $\tanh$ function; $\textbf{C}_s \in \mathbb{R}^{d\times d\times 2048}$, $\textbf{B}_s \in \mathbb{R}^{d\times d}$. The unbinding vector $\textbf{u}_t \in \mathbb{R}^{d}$ is given by,
\begin{equation}
 \textbf{u}_t = \textbf{U}\textbf{a}_t^{(u)}
\end{equation}
where $\textbf{U} \in \mathbb{R}^{d\times d}$ is a normalized Hadamard matrix. The ``filler vector" $\textbf{f}_t \in \mathbb{R}^{d} \rightarrow$  ``unbound" from the TPR representation $\textbf{S}_t$ with the ``unbinding vector" $\textbf{u}_t \rightarrow$ obtained by Equation \ref{eq:filler}.
\begin{equation} \label{eq:filler}
 \textbf{f}_t = \textbf{S}_t\textbf{u}_t
\end{equation}
LSTM is used to decode $\textbf{f}_t (t = 1,\ldots, n)$, where input is $\textbf{v}$ at $(t = 0)$, $\textbf{f}_t$ at time $t (t = 1,\ldots, n)$. 
In this work, we have demonstrated that if we decompose this vector $\textbf{f}_t$ in the recurrent neural network at time instance $t$, it can achieve considerable performance enhancement which is equivalent to the applying reinforcement learning to the SCN-LSTM$+$TPR network. This is the basis of our architecture and in the subsequent Sections, we will discuss the architecture in details.

\section{Our Architecture}
\label{sec:arc}
Context from visual images consists of a diverse range of embedded objects and when these are utilized for image captioning, a combination of regional or focused object gets highlighted as attention instead of the individuals. The main problem is lack of activation for the memory network for generation of likelihood of diverse activities and objects for composition of the sentence. However, if the visual feature is decomposed through heuristics (transformation based focus) and then placed in the model as generator or attention, it can help in better caption generation. In this section, we have featured different types of such decomposition techniques, their utility and effectiveness and their placement in memory networks. 

\subsection{Embedding + Hidden + TPR Architecture} \label{subsec:EHT}
This architecture is based on SCN-LSTM \cite{gan2017semantic} recurrent model with TPR \cite{smolensky1990tensor,palangi2017question,Sur2018Representation,huang2018tensor,huang2019tensor} attention where experiments were performed with different combinations of TPR $\textbf{T}$, Embedding $\textbf{x}_{t-1}$ and Hidden $\textbf{h}_{t-1}$ attention in SCN-LSTM. If Embedding and Hidden decomposition of SCN-LSTM are combined with TPR, the performance of SCN-LSTM gets enhanced as both the representation creates variations in the model and with time, the SCN-LSTM model has learned to be sensitive and react to the variation of the parameters. The decomposition of Embedding, Hidden along with estimated weights have learned to evolve meaningful information for the model networks. The main important part of the Embedding and Hidden decomposition is the constantly changing or shifting mode of operation, which unsaturated the long short term model memory and helped prevent appearance of similar kinds of word sequences as sentences. Mathematically, the equations for all decomposed model consist of the followings.
\begin{equation} \label{eq:emb}
 \textbf{x}_{*,t-1} = \textbf{W}_{x,*m} S \odot \textbf{W}_{x,*n} \textbf{x}_{t-1}
\end{equation}
\begin{equation} \label{eq:hid}
 \textbf{h}_{*,t-1} = \textbf{W}_{h,*m} S \odot \textbf{W}_{h,*n} \textbf{h}_{t-1}
\end{equation}
This architecture is the most primitive with $* = i/f/o/g$ . 
\begin{equation}
 \textbf{i}_t = \sigma(\textbf{W}_{xi}\textbf{x}_{i,t-1} + \textbf{W}_{hi}\textbf{h}_{i,t-1} + \textbf{W}_{Ti}\textbf{T} + \textbf{b}_{i})
\end{equation}
\begin{equation}
 \textbf{f}_t = \sigma(\textbf{W}_{xf}\textbf{x}_{f,t-1} + \textbf{W}_{hf}\textbf{h}_{f,t-1} + \textbf{W}_{Tf}\textbf{T} + \textbf{b}_{f})
\end{equation}
\begin{equation}
 \textbf{g}_t = \sigma(\textbf{W}_{xg}\textbf{x}_{g,t-1} + \textbf{W}_{hg}\textbf{h}_{g,t-1} + \textbf{W}_{Tg}\textbf{T} + \textbf{b}_{g})
\end{equation}
\begin{equation}
 \textbf{o}_t = \sigma(\textbf{W}_{xo}\textbf{x}_{o,t-1} + \textbf{W}_{ho}\textbf{h}_{o,t-1} + \textbf{W}_{To}\textbf{T} + \textbf{b}_{o})
\end{equation}
\begin{equation}
 \textbf{c}_{t} = \textbf{f}_{t} \odot \textbf{c}_{t-1} + \textbf{i}_{t} \odot \textbf{g}_{t} 
\end{equation}
\begin{equation}
 \textbf{h}_t = \textbf{o}_{t} \odot \tanh(\textbf{c}_{t})
\end{equation}
where we have $\textbf{i}, \textbf{f}, \textbf{g}, \textbf{o} \in \mathbb{R}^{m}$, $\textbf{b}_* \in \mathbb{R}^{m}$, $\textbf{W}_{x*} \in \mathbb{R}^{m\times d}$, $\textbf{W}_{h*} \in \mathbb{R}^{m\times m}$, $\textbf{W}_{T*} \in \mathbb{R}^{m\times d}$, $\textbf{W}_{h,*m} \in \mathbb{R}^{m\times 999}$, $\textbf{W}_{h,*n} \in \mathbb{R}^{m\times m}$, $\textbf{W}_{x,*m} \in \mathbb{R}^{m\times 999}$, $\textbf{W}_{x,*n} \in \mathbb{R}^{m\times d}$.
In Table \ref{table:table1}, Embedding and Hidden mean inclusion of Equations \ref{eq:emb} and Equations \ref{eq:hid} in place of traditional $\textbf{x}_{t-1}$ and $\textbf{h}_{t-1}$ in SCN-LSTM.  
where $\textbf{T} = \textbf{f}_t$ from Equation \ref{eq:filler}. This architecture is regarded as a TPR attention based SCN-LSTM and it had already out-performed the basic architecture for fusion of image features and the semantic tag features defined as SCN-LSTM in \cite{gan2017semantic}. In our model, we chained the knowledge and created structural composition and proper fusion of knowledge, which can be utilized and interpreted later.

\subsection{Embedding + Hidden + dTPR Architecture}
\begin{figure}[!h]
\centering
\includegraphics[width=.5\textwidth]{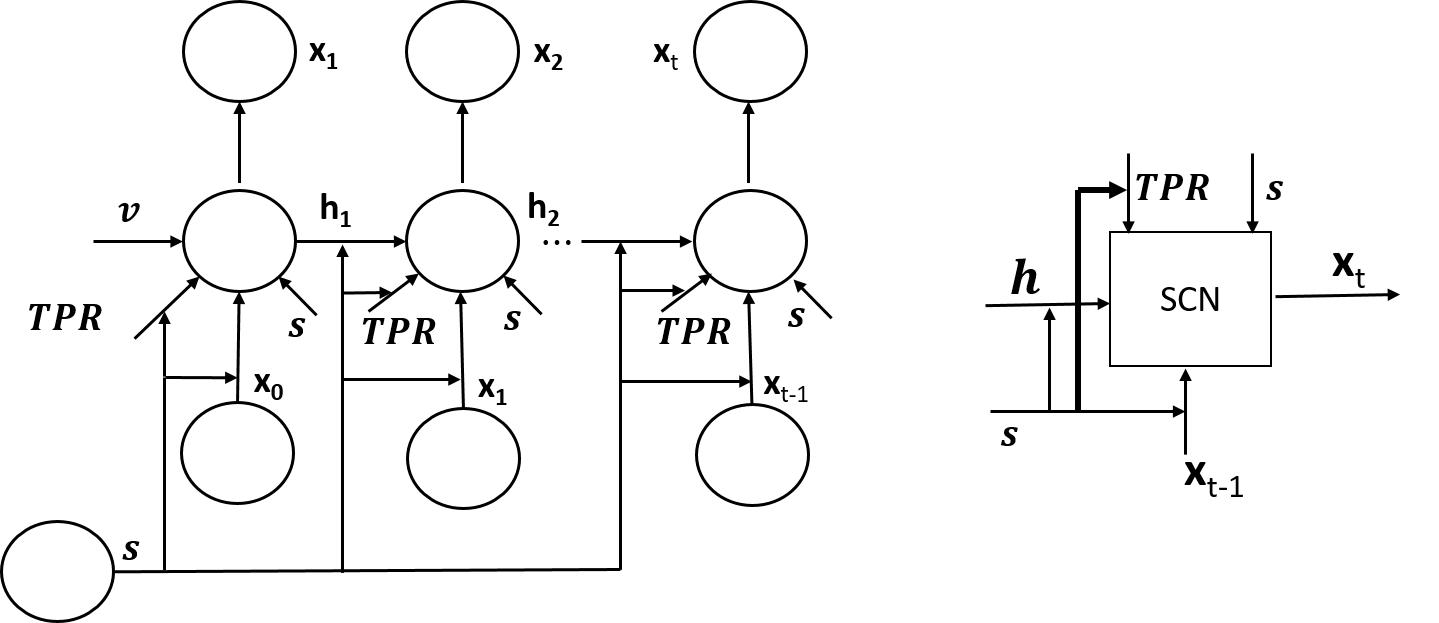}  
\caption{Overall Architecture with Decomposed TPR} \label{fig:dTPRarchitecture}
\end{figure}

In this work, we introduce the notion of decomposed TPR (dTPR) tensor and showed that certain combinations of this features with others can greatly enhance the performance. Here, we have Embedding and Hidden and dTPR, where all the components factorized for structure combination generation for captions. This architecture is very weighted architecture than the previous ones, but it performed much better and can be regarded as the state-of-the-art when only image features are concerned. Mathematically, the model can be represented as the following equations.
\begin{equation}
 \textbf{x}_{*,t-1} = \textbf{W}_{x,*m} S \odot \textbf{W}_{x,*n} \textbf{x}_{t-1}
\end{equation}
\begin{equation}
 \textbf{h}_{*,t-1} = \textbf{W}_{h,*m} S \odot \textbf{W}_{h,*n} \textbf{h}_{t-1}
\end{equation}
\begin{equation} \label{eq:dtpr}
 \textbf{T}_{*} = \textbf{P}_{*m} S \odot \textbf{P}_{*n} \textbf{T}
\end{equation}
These weighted parts are plugged into the memory network for enhancement of captions as the followings with added decomposed components $\textbf{T}_*$ with $* = i/f/o/g$. 
\begin{equation}
 \textbf{i}_t = \sigma(\textbf{W}_{xi}\textbf{x}_{i,t-1} + \textbf{W}_{hi}\textbf{h}_{i,t-1} + \textbf{W}_{Ti}\textbf{T}_{i} + \textbf{b}_{i})
\end{equation}
\begin{equation}
 \textbf{f}_t = \sigma(\textbf{W}_{xf}\textbf{x}_{f,t-1} + \textbf{W}_{hf}\textbf{h}_{f,t-1} + \textbf{W}_{Tf}\textbf{T}_{f} + \textbf{b}_{f})
\end{equation}
\begin{equation}
 \textbf{g}_t = \sigma(\textbf{W}_{xg}\textbf{x}_{g,t-1} + \textbf{W}_{hg}\textbf{h}_{g,t-1} + \textbf{W}_{Tg}\textbf{T}_{g} + \textbf{b}_{g})
\end{equation}
\begin{equation}
 \textbf{o}_t = \sigma(\textbf{W}_{xo}\textbf{x}_{o,t-1} + \textbf{W}_{ho}\textbf{h}_{o,t-1} + \textbf{W}_{To}\textbf{T}_{o} + \textbf{b}_{o})
\end{equation}
\begin{equation}
 \textbf{c}_{t} = \textbf{f}_{t} \odot \textbf{c}_{t-1} + \textbf{i}_{t} \odot \textbf{g}_{t} 
\end{equation}
\begin{equation}
 \textbf{h}_t = \textbf{o}_{t} \odot \tanh(\textbf{c}_{t})
\end{equation}
where $\textbf{P}_{*m} \in \mathbb{R}^{m\times 999}$, $\textbf{P}_{*n} \in \mathbb{R}^{m\times d}$ and the rest notations are same as Section \ref{subsec:EHT}. This architecture sees all the decomposition of hidden, embedding and TPR. Though it diversified the signatures of the sentences, it is not optimal with respect to the evaluated metrics. In our analysis, we found that Embedding + dTPR $(\textbf{x}_{*,t-1}, \textbf{h}_{t-1}, \textbf{T}_{*})$ setting provided maximum effectiveness though most of the architecture provided results that out-performed \cite{gan2017semantic}. Figure \ref{fig:dTPRarchitecture} provided an diagrammatic overview of the decomposed TPR based architecture, which is denoted by the extra set of equations (Equation \ref{eq:dtpr}) as $\textbf{T}_{*}$ to replace $\textbf{T}$ with $\textbf{f}_t$ from Equation \ref{eq:filler} in SCN-LSTM equation. However, Embedding + Hidden + dTPR $(\textbf{x}_{*,t-1}, \textbf{h}_{*,t-1}, \textbf{T}_{*})$ like settings overdo the decomposition and have lower performance. However, they are observed to get better with reinforcement learning. 

\section{Results \& Analysis}
\label{sec:res}

\begin{table*}[!h]
\centering
\caption{Performance Comparison \& Analysis for Different Image Captioning Architectures for Different Metrics}
\begin{tabular}{|c|c|c|c|c|c|c|c|c|}
\hline
 Algorithm & CIDEr-D & Bleu\_4 & Bleu\_3 & Bleu\_2 & Bleu\_1 & RG\_L & METEOR  &  SPICE \\ 
\hline \hline
    Adaptive \cite{lu2017knowing}  & 1.085 & 0.332 & 0.439 & 0.580 & 0.742 & -- & 0.266 & --  \\
    MSM \cite{yao2017boosting}  & 0.986 & 0.325 & 0.429 & 0.565 & 0.730 & -- & 0.251 & --  \\ 
    ERD \cite{yang1605encode}  & 0.895 & 0.298 & -- & -- & -- & -- & 0.240 & --  \\ 
    Att2in \cite{rennie2017self}  & 1.01 & 0.313 & -- & -- & -- & -- & 0.260 & --  \\  
    Up-Down \cite{anderson2018bottom}  & 105.4 & 0.334 & -- & -- & 0.745 & -- & 0.261 & 0.192  \\ %
    NBT \cite{lu2018neural}  & 1.07 & 0.347 & -- & -- & 0.755 & -- & 0.271 & 0.201  \\ 
    TPGN \cite{huang2018tensor}  & 0.909 & 0.305 & 0.406 & 0.539 & 0.709 & -- & 0.243 & --  \\ %
     ATPL \cite{huang2019tensor}  & 1.013 & 0.335 & 0.437 & 0.572 & 0.733 & -- & 0.258 & --  \\ %
    Attribute-Attention \cite{chen2018show} & 1.044 & 0.338 & 0.443 & 0.579 & 0.743 & 0.549  & -- & -- \\ %
    LSTM \cite{gan2017semantic} & 0.889 & 0.292 & 0.390 & 0.525 & 0.698 & -- & 0.238 & -- \\
    SCN \cite{gan2017semantic} & 1.012 & 0.330 & 0.433 & 0.566 & 0.728 & -- & 0.257 & --  \\ 
    \hline\hline
 
 (Embedding + TPR) & 1.001 & 0.332 & 0.437 & 0.573 & 0.736 & 0.543 & 0.256 & 0.184  \\ 
 
 (Hidden + TPR)  & 1.018 & 0.334 & 0.437 & 0.573 & 0.736 & 0.545 & 0.257 & 0.186  \\ 
 
 (Hidden + Embedding + TPR) \cite{Sur2018Representation} & 1.014 & 0.3352 & 0.439 & 0.575 & 0.737 & 0.546 & 0.257 & 0.1888  \\ 
 
 \textbf{(Embedding + dTPR)} & \textbf{1.022} & \textbf{0.338} & \textbf{0.443} & \textbf{0.578} & \textbf{0.737} & \textbf{0.546} & \textbf{0.256} & \textbf{0.1862}  \\ 
 
 (Hidden + dTPR)  & 1.0158 & 0.333 & 0.437 & 0.572 & 0.735 & 0.544 & 0.258 & 0.1864  \\ 
 
 (Hidden + Embedding + dTPR) & 1.006 & 0.331 & 0.435 & 0.570 & 0.734 & 0.542 & 0.256 & 0.186  \\ \hline \hline
  (Embedding + TPR) + RL$\dagger$ & 0.991 & 0.334 & 0.440 & 0.576 & 0.738 & 0.544 & 0.253 & 0.184  \\ 

  (Hidden + TPR)  + RL$\dagger$  & 1.001 & 0.335 & 0.440 & 0.576 & 0.738 & 0.547 & 0.255 & 0.184   \\ 
 
 \textbf{(Hidden + Embedding + TPR)  + RL}$\dagger$ & \textbf{1.002} & \textbf{0.338} & \textbf{0.442} & \textbf{0.578} & \textbf{0.738} & \textbf{0.546} & \textbf{0.255} & \textbf{0.185}  \\ 
 
  (Embedding + dTPR) + RL$\dagger$  & 1.003 & 0.336 & 0.441 & 0.577 & 0.738 & 0.545 & 0.255 & 0.184  \\ 
 
 (Hidden + dTPR)  + RL$\dagger$  & 0.998 & 0.336 & 0.439 & 0.573 & 0.734 & 0.543 & 0.254 & 0.184  \\ 
 
 (Hidden + Embedding + dTPR)  + RL$\dagger$ & 0.997 & 0.336 & 0.443 & 0.580 & 0.741 & 0.545 & 0.254 & 0.184  \\ \hline
 \multicolumn{3}{l}{\textsuperscript{$\dagger$}\footnotesize{Reinforcement Learning Used}} \\ 
\end{tabular}
\label{table:table1}
\end{table*}

\subsection{Dataset Preparation \& Training}
MSCOCO is used for experiments and analysis and this data is perhaps the most comprehensive data available. MSCOCO consists of 123K train images and 566K train sentences, where each image is associated with at least five sentences from a vocabulary of 8791 words with 5K images (with 25K sentences) for validation and 5K images (with 25K sentences) for testing \cite{gan2017semantic,Karpathy2014Deep}. Two set of image features being used: one is ResNet features \cite{he2016deep} $(\textbf{v})$ with 2048 dimension feature vector and another is Tag features $(S)$ with feature vector of 999 dimension \cite{gan2017semantic}. 
Tag contributes more when used with image features and the correlation based fusion has been the turning point for these image captioning application. The maximum is achieved in (tag + img) fusion combination and the result is provided in Table \ref{table:table1}. 
The initial state of $\textbf{c}_0$ and $\textbf{h}_0$ are initialized by through the image features as $\textbf{c}_0 = f_c(\textbf{v})$ and $\textbf{h}_0 = f_h(\textbf{v})$, where functions $f_c(.)$ and $f_h(.)$ are realized by two separate multilayer perceptrons (MLPs). Finally, SCN-LSTM decodes the likelihood of a word as  $(\textbf{h}_x)_t = \sigma_s(\textbf{W}_x\textbf{h}_t)$ where $(\textbf{h}_x)_t \in \mathbb{R}^V$, $\textbf{x}_t \in \mathbb{R}^d$ and we have $\textbf{x}_t = \textbf{W}_e (\textbf{h}_x)_t$, where $\sigma_s(.)$ is a softmax function; $\textbf{W}_x \in \mathbb{R}^{V \times 512}$. We have used the ResNet features $(\textbf{v})$ and Tag  features $(S)$ for each image from \cite{gan2017semantic}.

\subsection{Evaluation}
\begin{figure*}[!h] 
\centering 
\includegraphics[width=.75\textwidth]{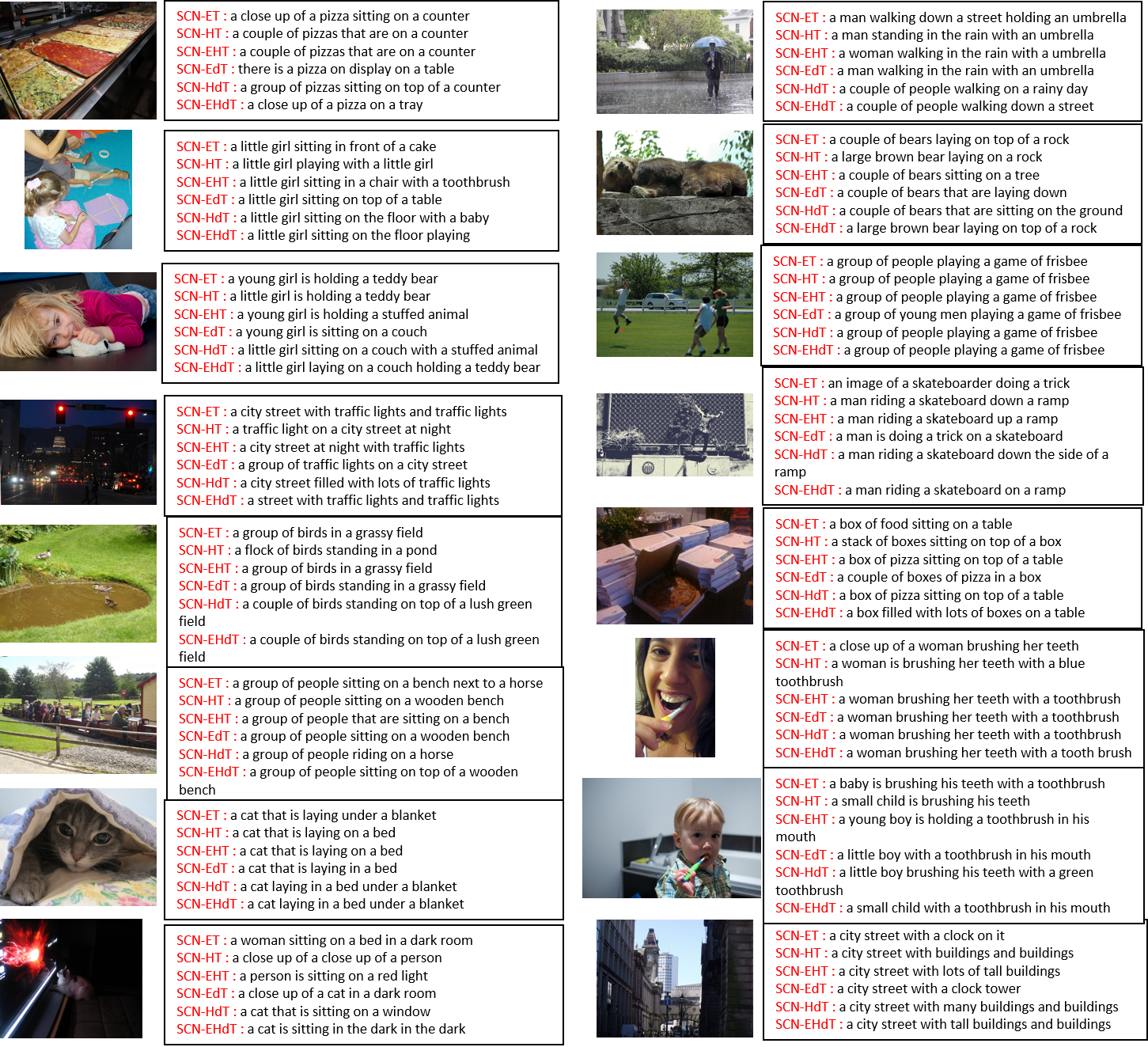}  
\caption{Qualitative Analysis.}
\label{fig:QualitativeAnalysis1}
\end{figure*}

Performance evaluation based on different language metrics like Bleu\_n $(n=1,2,3,4)$, METEOR, ROUGE\_L, CIDEr-D and SPICE are provided as these are standardized in the  community for image captioning research. However, none of the evaluation is complete and reflect very limited perspective of the generated captions. 
Table \ref{table:table1} provided a comparative study of our models with some of the existing works in this domain using these features and some works (like Attribute-Attention \cite{chen2018show}, NBT \cite{lu2018neural}) even used advanced features for their work and yet our work is comparable. Also, it would be injustice to compare models that have used other enhanced features and provided the improvements in image captioning. With only whole image set of features, this work can be regarded as the state-of-the-art performance with an effort to structure and characterize data features and generate longer and more descriptive captions. The main functional characteristics of our work is the decomposition of the generated/trained structural features and then composed new representation through different circumstances of weights and combinations and thus being able to derive better strategy for decoding and generation of the sentences. Most of our new architectures performed very well and either outperformed or at least same with the existing architectures, which do not have concrete reasoning behind their working principles.  However, SCN-LSTM + (Embedding + dTPR) is the clear winner (with respect to BLEU\_4) without Reinforcement Learning based training enhancement. With SCST reinforcement learning,  SCN-LSTM + (Hidden + Embedding + TPR) emerged as the best (with respect to BLEU\_4), establishing the fact that TPR based solutions are useful and can derive useful structural qualities for the system.
Though statistical evaluation metrics combination can provide many qualitative insights of the generated languages, they hardly reflected many aspects of languages including meaning, grammar, correct part-of-speech etc. These can only be evaluated through reading and hence we will provide comparison in terms of diversity and descriptive attributes. Figure \ref{fig:QualitativeAnalysis1} provided some examples and comparative instances, generated by different models.

\subsection{Reinforcement Learning Effects}
Self-critical Sequence Training (SCST) \cite{rennie2017self} based reinforcement learning has been used for gathering improvements in the performance of the models. 
Table \ref{table:table1} provided some evaluations of the architectures where SCST based reinforcement helped in improvement of the performance and achieved better BLEU\_4 for architectures. However, there are cases like (Embedding + dTPR), where the performance may get plunged due to RL over-fitting.
SCST utilized the gradient of the difference in performance between the generated and the referenced or baseline captions. In our case, SCST reinforcement learning is based on the gradient of CIDEr-D of the reference sentence and the intermediate generated caption with early stopping phenomenon. The batch-normalized CIDEr-D score is used as the gradient for updation of the prior weights of the models with a $\beta$ value of 0.7 with .7 and .3 weightage for CIDEr-D score based SCST and log-likelihood. It must be mentioned that SCST based reinforcement learning can promise improvement, but never guarantee improvement and this is demonstrated in Table \ref{table:table1}, mainly in case of saturating performance. However, more experiments can sometime provide improvement, which is not feasible without high-end GPUs and time.
SCST based reinforcement learning can be denoted with these equations,
\begin{equation} \label{eq:CR1}
 \frac{\delta L(\textbf{w})}{\delta \textbf{w}} = -\frac{1}{2b}\gamma \sum\limits_i \Phi(\textbf{y},\textbf{y}')
\end{equation}
\begin{equation} \label{eq:IR}
 \frac{\delta L(\textbf{w})}{\delta \textbf{w}} = -\frac{1}{2b}\gamma \sum\limits_i \Phi(\{y_1,\ldots,y_c\}, \{y'_1,\ldots,y'_{\grave{c}}\})
\end{equation}
where $\Phi(.)$ is the evaluation function or the reward function that evaluates certain aspects of the generated captions $\{y_1,\ldots,y_c\} \in \textbf{y}$ and the baseline captions $\{y'_1,\ldots,y'_{\grave{c}}\} \in \textbf{y}'$ with size $c$ and $\grave{c}$ respectively and $b$ is the mini-batch size considered.

\section{Discussion} \label{sec:discussion}
In this work, we discussed some improvements of the existing structures of feature decomposition and demonstrated that our approaches are better than previous works in all the possible metrics. We nurtured TPR and its interaction with other informative structures of the memory network and leveraged for variation generation and for identification of the attributes and interaction in images to appear in the sentences. Our mission is for a better representation and something that can be generalized for media features and help machines understand what is happening in the images. While, TPR succeeded in gathering improvement, we utilized different decomposition techniques for composition of structures which are as good as introducing reinforcement learning to some architectures. 
The future works can be concentrated on introducing more sophistication of the features and introduction of other useful components and structuring of the data that can differentiated by the models and generalize the representation to its unique sentence counterparts.

\bibliographystyle{IEEEbib}
\bibliography{icme2019template}

\end{document}